\documentclass[letterpaper, 10 pt, conference]{ieeeconf}  

\IEEEoverridecommandlockouts                              

\title{\LARGE \bf
SPLASH! Sample-efficient Preference-based inverse reinforcement learning for Long-horizon Adversarial tasks from Suboptimal Hierarchical demonstrations
}

\author{Peter Crowley$^{1*}$, Zachary Serlin$^2$, Tyler Paine$^{4,5}$, Makai Mann$^3$, Michael Benjamin$^{4}$, Calin Belta$^6$%
\thanks{
    $^{1}$Boston University, {\tt\small petertc@bu.edu}.
}%
\thanks{
    $^{2}$MIT Lincoln Laboratory, {\tt\small zachary.serlin@ll.mit.edu}.
}%
\thanks{
    $^{3}$MIT Lincoln Laboratory, {\tt\small makaim@cs.stanford.edu}.
}
\thanks{
    $^{4}$MIT Pavlab, {\tt\small \{tpaine,mikerb\}@mit.edu}.
}%
\thanks{$^{5}$Woods Hole Oceanographic Institution.}%
\thanks{$^{6}$University of Maryland, College Park, {\tt\small calin@umd.edu}.}
}%

\usepackage{makeidx}      
\usepackage{multicol}        
\usepackage{multirow}
\usepackage[bottom]{footmisc} 

\usepackage{graphicx}        

\usepackage[font=footnotesize]{caption}
\usepackage[font=footnotesize]{subcaption}

\usepackage[dvipsnames]{xcolor}

\usepackage{algorithm}
\usepackage[noend]{algpseudocode}

\usepackage[cmex10]{amsmath}
\usepackage{amssymb}

\usepackage{amsthm}
\usepackage{setspace}

\usepackage[hidelinks]{hyperref}

\theoremstyle{plain}

\begin{document}

\maketitle
\thispagestyle{empty}
\pagestyle{empty}

\begin{abstract}
Inverse Reinforcement Learning (IRL) presents a powerful paradigm for learning complex robotic tasks from human demonstrations. However, most approaches make the assumption that expert demonstrations are available, which is often not the case. Those that allow for suboptimality in the demonstrations are not designed for long-horizon goals or adversarial tasks. Many desirable robot capabilities fall into one or both of these categories, thus highlighting a critical shortcoming in the ability of IRL to produce field-ready robotic agents. We introduce Sample-efficient Preference-based inverse reinforcement learning for Long-horizon Adversarial tasks from Suboptimal Hierarchical demonstrations (SPLASH), which advances the state-of-the-art in learning from suboptimal demonstrations to long-horizon and adversarial settings. We empirically validate SPLASH on a maritime capture-the-flag task in simulation, and demonstrate real-world applicability with sim-to-real translation experiments on autonomous unmanned surface vehicles. We show that our proposed methods allow SPLASH to significantly outperform the state-of-the-art in reward learning from suboptimal demonstrations.
\end{abstract}

\section{Introduction} \label{sec:introduction}
Ever since Mnih et al. \cite{mnih2015human} first used deep neural networks as Q-function approximators for reinforcement learning (RL) \cite{sutton2018reinforcement}, deep RL (DRL) algorithms have achieved superhuman performance on tasks with high-dimensional state and observation spaces, such as video games \cite{schaul2015prioritized, van2016deep, wang2016dueling, hessel2018rainbow, badia2020agent57, kapturowski2022human}. Despite this, DRL has been underutilized in the development of real-world autonomous systems. This untapped potential is partially due to the challenge of designing reward functions for long-horizon and adversarial tasks. Sparse reward functions may accurately represent the overall goal of a task, but 
RL agents training on sparse rewards will often struggle to achieve goals through random exploration \cite{andrychowicz2017hindsight}, especially in the case of long-horizon tasks. While reward shaping can alleviate this issue, handcrafted reward functions may not produce desirable strategies and can instead lead to unintended behaviors that are detrimental to the overall goal \cite{ng1999policy, amodei2016concrete}.

\begin{figure}[t]
    \centering
    \includegraphics[width=\linewidth]{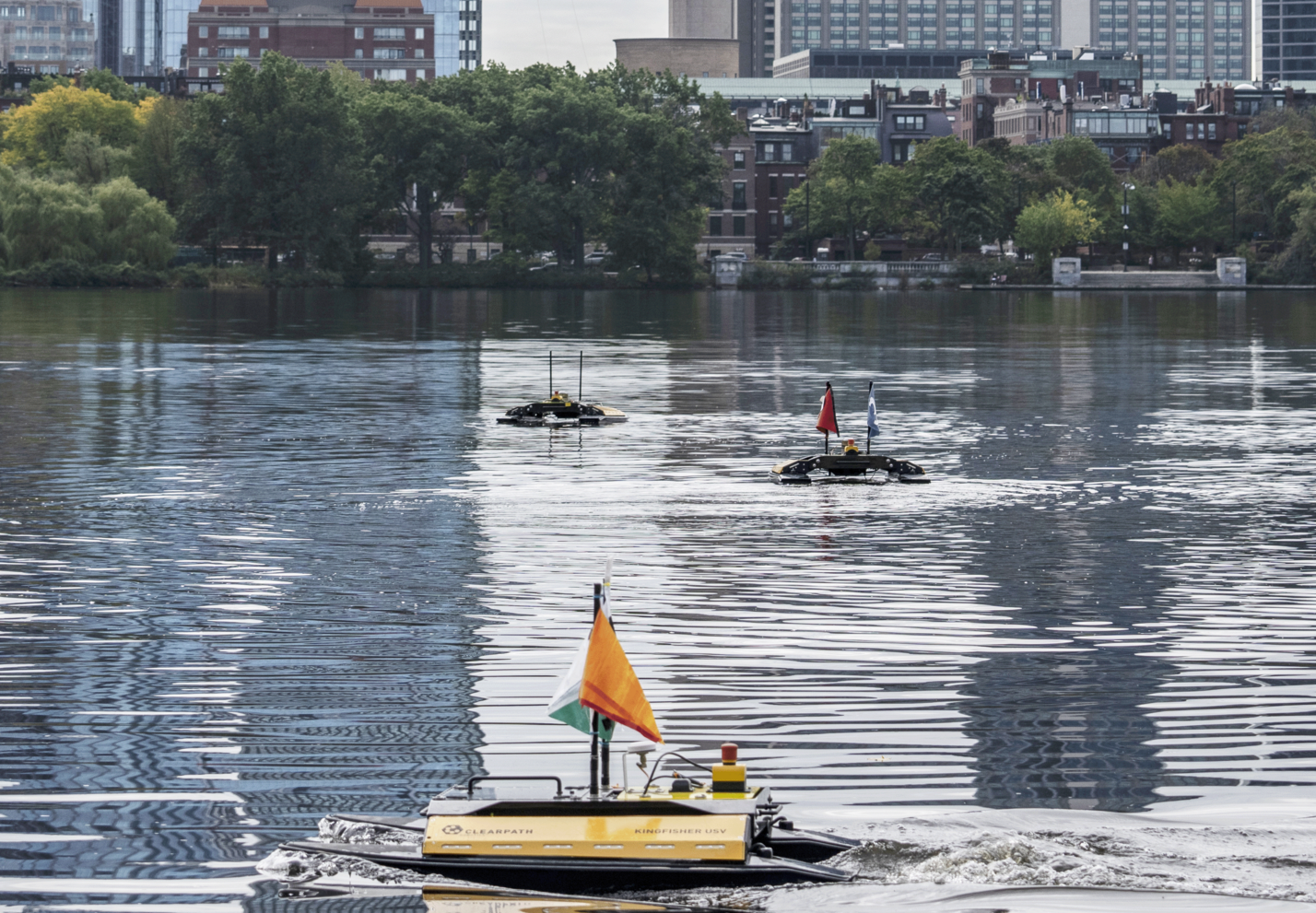}
    \caption{
        Autonomous Heron surface vehicles playing capture-the-flag on the Charles River in Boston using policies derived from our SPLASH framework.
    }
    \label{fig:1}
    \vspace{-4mm}
\end{figure}

As an alternative to reward-shaping by hand, IRL uses human demonstrations to learn a reward function corresponding to the goal of the demonstrated task \cite{ng2000algorithms, arora2021survey}. DRL can then be used to train agents on the learned reward function. However, many IRL algorithms require expert demonstrations \cite{ziebart2008maximum, boularias2011relative, finn2016guided} which limits their real-world applicability as human demonstrations are often highly suboptimal. In contrast, Preference-based IRL (PBIRL) methods \cite{sugiyama2012preference, wirth2016model, ibarz2018reward, brown2019extrapolating, palan2019learning} do not require expert demonstrations. Instead, they recover a reward function from human-ranked or performance-characterized demonstrations. Ideally, the learned reward function will reflect the underlying intentions of the demonstrator and accurately extrapolate beyond suboptimal demonstrator performance levels. While some PBIRL algorithms are able to learn from suboptimal demonstrations \cite{brown2019extrapolating, palan2019learning}, these methods struggle to learn accurate reward functions for long-horizon and hierarchical games, such as Montezuma's Revenge or H.E.R.O. \cite{brown2019extrapolating}. While later works make improvements to the process of learning from suboptimal demonstrations \cite{brown2020better, chen2021learning}, such as automatic demonstration ranking, they do not focus on how to learn reward functions for long-horizon or adversarial tasks.

To address this shortcoming, we propose a novel IRL algorithm, called Sample-efficient Preference-based inverse reinforcement learning for Long-horizon Adversarial tasks from Suboptimal Hierarchical demonstrations (SPLASH), which extends the state-of-the-art in learning from suboptimal demonstrations to long-horizon and adversarial reward learning. Specifically, SPLASH is a PBIRL algorithm that: \textbf{(1)} downsamples full trajectories of long-horizon tasks for efficient pairwise comparison, \textbf{(2)} checks for alignment between demonstration ranking and the available success/score criteria of the task, removing contradictions from the training data, \textbf{(3)} selects for reward functions that increase with progress and decrease with losses in adversarial settings, and \textbf{(4)} smooths the learned reward function with temporal consistency regularization. Additionally, we collect demonstrations at the options level \cite{sutton1999between}, rather than low-level actions, for more sample-efficient modeling of demonstrator policies with behavioral cloning.

We evaluate our approach in an open-source capture-the-flag (CTF) Gymnasium environment \cite{pyquaticus} and demonstrate the sim-to-real translatability of a learned options-level policy on a maritime CTF testing platform \cite{moos-ivp} as shown in Fig. \ref{fig:1}. Our experiments show that, when given suboptimal demonstrations of CTF, SPLASH is able to learn reward functions that accurately reflect the performance and progress of CTF agents, whereas the state-of-the-art in PBIRL struggles to infer the underlying intentions and goals of the game. To the best of our knowledge, SPLASH is the first inverse reinforcement learning approach designed to learn from suboptimal demonstrations of long-horizon and adversarial tasks.

\section{Related Work}
\subsection{Learning from Demonstrations}
Algorithms designed to learn from demonstrations can generally be classified as either imitation learning or IRL. The primary distinction between the two is that imitation learning directly learns a model of the demonstrator policy while IRL learns a reward function that reflects the demonstrator's intentions. This reward function can be used by any reinforcement learning algorithm to learn the demonstrator policy. As opposed to imitation learning, IRL can provide an avenue for producing an optimal policy from suboptimal demonstrations by recovering a demonstrator's true intentions in the learned reward function.

\subsection{Imitation Learning}
The most straightforward imitation learning approaches use behavioral cloning (BC) to directly mimic the demonstrator's policy with supervised learning \cite{pomerleau1991efficient, ijcai2018p687}. Because BC trains only on the experiences seen in the demonstrations, it struggles to generalize well when the demonstrations lack sufficient coverage of the state space. The state-of-the art in imitation learning methods frame the problem from within the context of generative adversarial networks (GANs) \cite{goodfellow2014generative} by modeling the demonstrator action distribution with the generator, and using the discriminator as a proxy for the task's cost-function. However, GANs often struggle to converge, and both imitaiton learning and IRL approaches that utilize GANs \cite{ho2016generative, fu2017learning} tend to perform poorly on complex high-dimensional tasks such as Atari \cite{tucker2018inverse, brown2019extrapolating, brown2020better}. For these reasons, we utilize BC for learning approximations of the demonstrator policy in our approach. We address BC's issues of generalization and sample-efficiency by collecting subtask-labeled hierarchical demonstrations.

\subsection{Preference-Based IRL}
Preference-based IRL (PBIRL) methods \cite{sugiyama2012preference, wirth2016model, ibarz2018reward, brown2019extrapolating, palan2019learning} learn a reward function based on ranked demonstrations. Unlike maximum entropy IRL, PBIRL does not assume that all demonstrations are equally suitable for the given task. In theory, this relaxation should allow PBIRL to recover the true intentions of a suboptimal demonstrator. However, a major drawback to these methods is that they often require many human hours of manual demonstration ranking to obtain good results.

Brown et al. \cite{brown2019extrapolating} were the first to study in depth how PBIRL could be used to outperform highly suboptimal demonstrators. Their algorithm, Trajectory-ranked Reward EXtrapolation (T-REX), is able to learn reward functions capable of accurately extrapolating beyond demonstrator performance levels. However, T-REX struggles to learn good reward functions for long-horizon Atari games such as Montezuma’s Revenge and Hero. This shortcoming can likely be attributed to the inherent difficulty of reward learning for a complex task, and the lack of comparability between different sections of a long-horizon task. In other words, it can be difficult for a human to determine the relative ranking of two trajectories when such trajectories demonstrate different subtasks in the overall task \cite{wirth2017survey}.

Rather than relying on humans for demonstration ranking, which can be costly and time-consuming, Disturbance-based Reward EXtrapolation (D-REX) automatically generates a set of ranked demonstrations by injecting noise into the demonstrator policy \cite{brown2020better}. From there, T-REX is used to recover the reward function. Brown et al. show that D-REX is able to significantly outperform suboptimal demonstrators on both continuous control and Atari tasks. While Chen et al. \cite{chen2021learning} propose a method to more rigorously characterize the task-specific relationship between noise and performance degradation, their approach utilizes an initial adversarial step \cite{fu2017learning}, which is computationally expensive to run and tends to perform poorly on high-dimensional problems \cite{tucker2018inverse, brown2019extrapolating, brown2020better}. Additionally, they assume that the noise-performance degradation curve can be fit to a user-specified function (such as sigmoid), which may introduce bias to the algorithm's perception of trajectory return bounds. Because of this, we use D-REX as the key inspiration and benchmark for our approach to learning from suboptimal demonstrations.

\section{Preliminaries}
\subsection{Markov Decision Processes}
Formally, an MDP is defined as a tuple $\langle \mathcal{S}, \mathcal{A}, T, R, \gamma \rangle$ where $\mathcal{S}$ is the state space, $\mathcal{A}$ is the action space, $T : \mathcal{S} \times \mathcal{A} \times \mathcal{S} \to [0, 1]$ is the transition function, $R : \mathcal{S} \times A \to \mathbb{R}$ is the reward function, and $\gamma\in\mathbb{R}$ is the discount factor. A policy in an MDP maps the current state to a probability distribution over the action space $\pi : S \times \mathcal{A} \to [0,1]$. The value of a state $s \in S$ under policy $\pi$ is given as the discounted sum of rewards received by following policy $\pi$ from $s$: $V_\pi(s) = \mathbb{E}_\pi[\sum_{t=0}^\infty \gamma^t R(s_t, a_t)]$, where $a_t$ is the action at time $t$ and $s_0=s$. For a given MDP, a policy is considered optimal if it maximizes the cumulative discounted reward an agent receives from the environment: $\pi^* =\text{argmax}_\pi V_\pi(s) \; \forall s \in \mathcal{S}$.

\subsection{The Options Framework}
The options framework abstracts the decision-making process of hierarchical RL problems to the selection of high-level behaviors or subtasks (options) \cite{sutton1999between}. Formally, an option $\omega$ is defined by a triplet $\langle I_\omega, \pi_\omega, \beta_\omega \rangle$ where $\pi_\omega$ is the intra-option policy, $I_\omega \subseteq{\mathcal{S}}$ is the set of states at which $\pi_\omega$ can be used, and $\beta_\omega : \mathcal{S} \to [0,1]$ is a termination function that defines the probability of switching from option $\omega$ to a new option at a given state. The policy-over-options $\pi_\Omega : \mathcal{S} \times \Omega \to [0,1]$ defines a discrete probability distribution over option selection conditioned on the state, where $\Omega$ is the set of all options.

\subsection{Disturbance-based Reward Extrapolation}
Disturbance-based Reward EXtrapolation (D-REX) \cite{brown2020better} is a preference-based IRL (PBIRL) algorithm. In general, PBIRL deals with pairwise rankings of trajectories, where a trajectory $\tau$ is defined as a finite sequence of state-action pairs sampled from the demonstrations: $\{(s_t, a_t), (s_{t+1}, a_{t+1}), (s_{t+2}, a_{t+2}), ...\}$. The goal of PBIRL is to learn a reward function $R_\theta(s)$ such that $\sum_{s \in \tau_i} R_\theta(s) < \sum_{s \in \tau_j} R_\theta(s)$ is satisfied for all pairs of equal-length trajectories $(\tau_i \prec \tau_j)$, where $\tau_i \prec \tau_j$ indicates that $\tau_j$ is preferential to $\tau_i$. The D-REX objective is to to minimize the following binary cross-entropy loss function:

\vspace{-4mm}
\begin{equation}\label{eq:1}
\mathcal{L}(\theta) = -\log{\frac{\exp{\sum_{s \in \tau_j} R_\theta(s)}}{\exp{\sum_{s \in \tau_i} R_\theta(s)} + \exp{\sum_{s \in \tau_j} R_\theta(s)}}}
\end{equation}

\noindent
for all pairs $(\tau_i,\tau_j)$ where $\tau_i \prec \tau_j$.

To generate a training dataset of preference-labeled trajectory pairs, D-REX uses behavioral cloning to learn an approximation of the demonstrator policy $\pi_\text{BC}$, and then rolls out the policy according to an $\epsilon$-greedy noise schedule $\mathcal{E} = (0.01, 0.25, 0.5, 0.75, 1.0)$:

\vspace{-2mm}
\begin{equation}\label{eq:2}
\pi_\epsilon(a|s,\epsilon) = \left\{
\begin{array}{ll}
    1-\epsilon + \frac{\epsilon}{|\mathcal{A}|}, 
    & \text{if} \; \pi_{\text{BC}}(s) = a\\
    \frac{\epsilon}{|\mathcal{A}|},
    & \text{otherwise}\\
\end{array} \right..
\end{equation}

\noindent
This generates trajectories with varying levels of noise injection $\epsilon \in \mathcal{E}$. The key assumption is that trajectories decrease in quality with increased noise. For example: $\epsilon_{\tau_1} > \epsilon_{\tau_2} \implies \tau_1 \prec \tau_2$. Additionally, a set of no-op trajectories is generated and placed at the lowest rank to select for reward functions that value action over doing nothing. In this way, a set of ranked trajectories can be automatically generated.

In order to augment the training data, D-REX uses partial trajectories or ``snippets" $\tilde{\tau}$ subsampled from the full rollouts $\tau$ to construct the dataset of trajectory pairs. D-REX only selects partial trajectory pairs $\tilde{\tau}_i \prec \tilde{\tau}_j$ that satisfy $t_{i_0} \leq t_{j_0}$ where $t_{i_0}$ and $t_{j_0}$ are the starting time steps of $\tilde{\tau}_i$ and $\tilde{\tau}_j$ within $\tau_i$ and $\tau_j$ respectively. This helps prevent comparisons between earlier sections of higher-ranked rollouts and later parts of lower-ranked rollouts. If this was not done, low-value states visited early in a good rollout may be viewed as better than high-value states from later sections of a bad rollout. Brown et al. \cite{brown2019extrapolating} note that this requirement will select for reward functions that increase monotonically with progress.

\section{SPLASH}
SPLASH utilizes the basic framework of PBIRL to learn reward functions: \textbf{(a)} obtain a set of ranked demonstrations, and \textbf{(b)} use a preference objective, such as \eqref{eq:1}, to learn a reward function. This section outlines and motivates our approach, which involves the novel synthesis of the following elements: options-level demonstrations and behavioral cloning, downsampled full trajectory pairs for preference comparison, success and progress-based learning constraints, and temporal consistency regularization for the reward function. Our implementation of SPLASH provides the choice to use or omit each of these elements depending on its necessity for the task.

\subsection{Options-Level Demonstrations}
As in D-REX, we first collect a set of human demonstrations of the task with which to model the demonstrator policy. However, instead of collecting demonstrations as sequences of state and low-level action pairs, we collect them at the options level and use a mixture-of-options formulation \cite{henderson2018optiongan, daniel2012hierarchical} to model the demonstrator policy: $\pi(a|s) = \sum_{\omega \in \Omega} \pi_\Omega(\omega|s) \pi_\omega(a|s)$. Here, $\pi_\omega$ is an intra-option policy and $\pi_\Omega$ is the policy-over-options. This formulation uses $\pi_\Omega$ as a gating function to control option activation. We pre-define the intra-option policies $\pi_\omega$ with behavioral primitives which allows us to collect demonstrations at the options level. At every timestep the demonstrators will select an option $\omega \in \Omega$, such as ``attack" or ``defend", instead of a low-level action such as ``drive forward". Therefore, trajectories in the set of human demonstrations $\mathcal{D}_\Omega$ will be sequences of state-option pairs: $\{(s_t, \omega_t), (s_{t+1}, \omega_{t+1}), (s_{t+2}, \omega_{t+2}), ...\}$.

Similar to D-REX, SPLASH uses BC to model the demonstrator policy. However, because this policy is trained on state-option trajectories, instead of state-action, SPLASH produces an approximation of the demonstrator's policy-over-options $\pi_{\Omega_{\text{BC}}}(\omega|s)$, rather than $\pi_{\text{BC}}(a|s)$ which D-REX produces. As shown in the results, we find that collecting options-level demonstrations for behavioral cloning is more sample efficient for producing a policy that accurately reflects the overall demonstrator strategy in a complex long-horizon task, such as CTF. Additionally, using pre-defined intra-option policies can produce more predictable and interpretable behaviors.

\subsection{Downsampled Full Trajectories}
As in many preference-based learning approaches \cite{christiano2017deep, ibarz2018reward, palan2019learning, brown2019extrapolating, brown2020better}, we use \eqref{eq:1} as our main objective for training the reward function. This loss function requires a set of pairwise-ranked trajectories $\mathcal{P} = \{(i,j) : \tau_i \prec \tau_j \}$, which we generate automatically using D-REX's approach of rolling out the BC demonstrator policy with noise injection:

\vspace{-2mm}
\begin{equation}\label{eq:4}
\pi(a|s, \epsilon) = \sum_{\omega \in \Omega} \pi_{\Omega, \epsilon}(\omega|s, \epsilon) \pi_\omega(a|s)
\end{equation}

\begin{equation}\label{eq:5}
\pi_{\Omega, \epsilon}(\omega|s, \epsilon) = \left\{
\begin{array}{ll}
    1-\epsilon + \frac{\epsilon}{|\Omega|}, 
    & \text{if} \; \pi_{\Omega_\text{BC}}(s) = a\\
    \frac{\epsilon}{|\Omega|},
    & \text{otherwise}\\
\end{array} \right..
\end{equation}

\noindent
Rather than sub-sampling snippets of these rollouts for partial trajectory comparisons as is done in D-REX, SPLASH compares full trajectories (rollouts) of equal length. Many tasks that D-REX was tested on, such as HalfCheetah or Hopper, involve constant repetition of a motion or strategy that a partial trajectory $\tilde{\tau}$ can fully capture many times over. However, for long-horizon tasks such as CTF, a full trajectory that demonstrates the complete attempted execution of a task will illustrate the overall objective better than a partial snippet of such a trajectory. Additionally, comparing full trajectories of a long-horizon task avoids making illogical comparisons between partial trajectories that demonstrate different subtasks in the overall task \cite{wirth2017survey}. For these reasons, SPLASH constructs the dataset on which to train \eqref{eq:1} with pairs full trajectories that are downsampled for computational efficiency. While D-REX does employ trajectory downsampling for Atari games by only keeping every fourth frame, this amounts to eliminating redundant information as every Atari observation consists of 4 consecutive frames. We use much higher downsampling rates to account for full trajectories up to 50$\times$ longer than the partial trajectories used by D-REX, while also avoiding rates too high to capture the overall behavior exhibited by agents.

Because SPLASH cannot augment the training dataset by comparing trajectory snippets, each rollout in each noise bin $\epsilon$ is paired with each rollout in every other noise bin to create the largest possible set of trajectory pairs. SPLASH uses D-REX's noise-based ranking rule to assign preference labels to the pairs. However, to account for the imperfect assumptions of this approach, SPLASH leverages a task's scoring metric and success criterion to prune the training dataset of trajectory pairs whose preference labels are inconsistent with these factors. We represent the user-defined score function for a trajectory as $\eta(\tau)$ and define the trajectory success function as:

\vspace{-3mm}
\begin{equation}\label{eq:5}
\psi(\tau) = \left\{
\begin{array}{rl}
    1, & \text{if } \tau \text{ is a success} \\
    -1, & \text{if } \tau \text{ is a failure} \\
    0, & \text{otherwise}\\
\end{array} \right.
\end{equation}

\noindent
where failure and success conditions can be defined for the specific task. We include an option for neither success nor failure to account for tasks, such as adversarial games, in which a net neutral outcome is possible, as well as infinite horizon tasks in which there is not a clearly defined termination condition. To qualify for the training dataset, SPLASH requires that a trajectory pair $\tau_i \prec \tau_j$ satisfies $\psi(\tau_i) \leq \psi(\tau_j)$ and $\eta(\tau_i) \leq \eta(\tau_j)$. That is, the higher-ranked trajectory must have a score and degree of success greater than or equal to that of the lower-ranked trajectory.

While the score and successfulness of demonstrations may not be given initially, it is often easy to define a success criterion and scoring metric with which to label the demonstrations after they have been generated. Additionally, there is precedent for using success and failure-labeled trajectories to learn from suboptimal demonstrations \cite{shiarlis2016inverse}. Therefore, we view success and score-based trajectory pair pruning as a useful and widely applicable method. We note that using full trajectory pairs makes this approach more viable than it would have been for D-REX. This is because the entirety of a demonstration's score and successfulness cannot necessarily be attributed to a partial snippet of the full trajectory.

\subsection{Initial-Final State Comparisons}
To learn reward functions that capture the zero-sum nature of adversarial settings, we consider the win, lose, or draw nature of such tasks in the context of progress, i.e., from the start of a game, agents can gain or loose progress towards the ultimate goal of winning. The implication of this for SPLASH is that the reward at the final state of a winning demonstration should be greater than that of any initial state, and conversely the reward at the final state of a losing demonstration should be less than that of any initial state. SPLASH enforces these constraints for each trajectory pair $(\tau_i, \tau_j)$ with a state-preference objective similar to \eqref{eq:1}:

\vspace{-4mm}
\begin{equation}\label{eq:6}
\mathcal{L}_\text{IF}(\theta) = \phi(\tau_i, \tau_j) + \phi(\tau_j, \tau_i) + \phi(\tau_i, \tau_i) + \phi(\tau_j, \tau_j)
\end{equation}

\vspace{-5mm}
\begin{equation}\label{eq:7}
\phi(\tau_1, \tau_2) =
-\psi(\tau_2)
\log{
    \frac{
        e^{R_\theta\big(\tau_2(t_f)\big)}
    }
        {
        e^{R_\theta\big(\tau_1(t_0)\big)} + e^{R_\theta\big(\tau_2(t_f)\big)}
    }
}
\end{equation}

\noindent
where $\tau(t_0)$ is the initial state of a trajectory, and $\tau(t_f)$ is the final state. With this additional loss term, the SPLASH objective becomes $\mathcal{L} = \mathcal{L}_\text{PBIRL} + \lambda_\text{IF}\mathcal{L}_\text{IF}$ where $\mathcal{L}_\text{PBIRL}$ is the main PBIRL objective \eqref{eq:1}, and $\lambda_\text{IF}$ is a regularization coefficient. In this way, we select for reward functions that increase with progress, decrease when progress is lost, remain above the initial reward when the advantage is possessed, and fall below the initial reward when the advantage is lost.

In contrast, the unidirectional notion of progress built into D-REX's time-based snippet selection rule ($t_{i_0} \leq t_{j_0}$) prevents D-REX from learning reward functions that can fall below the initial reward. This rule will build pairs of partial trajectories in which the better trajectory $\tilde{\tau}_j$ occurs later than the worse trajectory $\tilde{\tau}_i$ and therefore the earliest segments will function as a lower bound on state-based reward. This prevents D-REX from accurately capturing rewards for adversarial games and may cause problems when comparing partial trajectories from demonstrations that end in a loss. For example, at two high noise levels $0 \ll \epsilon_1 < \epsilon_2 < 1$, a trajectory pair ($\tau_{\epsilon_1}, \tau_{\epsilon_2}$) may be produced such that both trajectories end in a loss where $\psi(\tau_{\epsilon_1}) = \psi(\tau_{\epsilon_2}) = -1$ and $\eta(\tau_{\epsilon_2}) < \eta(\tau_{\epsilon_1}) < 0$. According to D-REX's noise-based ranking and time-based snippet selection rules, snippets from the very end of $\tau_{\epsilon_1}$, where the score is low, are considered preferential to those at the beginning of $\tau_{\epsilon_2}$, where the score may not be low yet. SPLASH avoids this potential contradiction with the score both by constraining that $\big\{R_\theta\big(\tau_{\epsilon_1}(t_0)\big), R_\theta\big(\tau_{\epsilon_2}(t_0)\big)\big\} > \big\{R_\theta\big(\tau_{\epsilon_1}(t_f)\big), R_\theta\big(\tau_{\epsilon_2}(t_f)\big)\big\}$ with \eqref{eq:6}, and by comparing the full trajectories, instead of snippets, for which D-REX's noise-based preference label $\tau_{\epsilon_1} \succ \tau_{\epsilon_2}$ is valid because $\eta(\tau_{\epsilon_1}) > \eta(\tau_{\epsilon_2})$.  

\subsection{Temporal Consistency Reward Regularization}
To produce smooth reward functions well-suited for RL, we regularize by the first and second order derivatives of the reward function. We incorporate this temporal consistency regularization because training on highly downsampled trajectories may lose information about intermediate states that are not sampled and therefore their corresponding rewards cannot be tuned by $\mathcal{L}_\text{PBIRL}$. SPLASH approximates these derivatives with reward differentials on each trajectory in a trajectory pair $(\tau_i, \tau_j)$:

\vspace{-3mm}
\begin{equation}\label{eq:8}
\mathcal{L}_{dR}(\theta) =
\lambda_1 \frac{
    \sum_{\{\tau_i, \tau_j\}}{
        \mathcal{L}_{d1}(\tau)
    }
}
{
    |\tau_j| + |\tau_j| - 2
} +
\lambda_2 \frac{
    \sum_{\{\tau_i, \tau_j\}}{
        \mathcal{L}_{d2}(\tau)
    }
}
{
    |\tau_j| + |\tau_j| - 4
}
\end{equation}

\begin{equation}\label{eq:9}
\mathcal{L}_{d1}(\tau) =
\sum_{\substack{s_t \in \tau \\ t=0}}^{|\tau|-2}{
    \big|
        R_\theta(s_{t+1}) - R_\theta(s_t)
    \big|
}
\end{equation}

\vspace{-1mm}
\begin{equation}\label{eq:10}
\mathcal{L}_{d2}(\tau) =
\sum_{\substack{s_t \in \tau \\ t=0}}^{|\tau|-3}{
    \big|
        R_\theta(s_{t+2}) - 2R_\theta(s_{t+1}) + R_\theta(s_t)
    \big|
}
\end{equation}

\noindent
where $\lambda_1$ and $\lambda_2$ are the first and second order reward derivative regularization coefficients, and $|\tau|$ is the number of states in trajectory $\tau$. With this final addition, we arrive at the full SPLASH loss function and present the complete procedure in algorithm~\ref{alg:SPLASH}:

\begin{equation}\label{eq:11}
\mathcal{L}(\theta) =
\mathcal{L}_\text{PBIRL}(\theta) +
\lambda_\text{IF}\mathcal{L}_\text{IF}(\theta) +
\mathcal{L}_{dR}(\theta).
\end{equation}

\begin{algorithm}[t]
\caption{SPLASH}
\label{alg:SPLASH}
\begin{algorithmic}[1]
    \State {\bfseries Input:} Options-level demonstrations $\mathcal{D}_\Omega$, noise schedule $\mathcal{E}$, number of rollouts $M$, score and success functions $\eta$ and $\psi$, regularization coefficients $\lambda_1$, $\lambda_2$, and $\lambda_\text{IF}$
    \State Run behavioral cloning on $\mathcal{D}_\Omega$ to obtain $\pi_{\Omega_\text{BC}}$
   \For{$\epsilon_i \in \mathcal{E}$}
        \State Generate $M$ trajectories with $\pi_{\text{BC}}(\cdot| \epsilon_i)$ \eqref{eq:4}
   \EndFor
   \State Initialize trajectory pair dataset $\mathcal{P} = \{(i,j) : \tau_i \prec \tau_j \}$
   \For{$\epsilon_i \in \mathcal{E}$}
        \For{$\epsilon_j \in \mathcal{E}$ \textbf{st} $\epsilon_i < \epsilon_j$}
           \For{$\tau_i \sim \pi_{\text{BC}}(\cdot| \epsilon_i)$}
                \For{$\tau_j \sim \pi_{\text{BC}}(\cdot| \epsilon_j)$}
                    \If{
                        $\psi(\tau_i) \leq \psi(\tau_j) \land \eta(\tau_i) \leq \eta(\tau_j)$
                    }
                        \State Add $(\tau_i \prec \tau_j)$ to $\mathcal{P}$
                    \EndIf
                \EndFor
           \EndFor
        \EndFor
   \EndFor
   \State Initialize reward function $R$ with weights $\theta$
   \State Optimize $R_\theta$ according to \eqref{eq:11} on all $(\tau_i, \tau_j) \in \mathcal{P}$
\end{algorithmic}
\end{algorithm}

\section{Experiments and Results} \label{sec:results}
\subsection{Maritime Capture-the-Flag Task}
We experimentally validate SPLASH in a maritime capture-the-flag Gymnasium environment \cite{pyquaticus}. As opposed to many of the standard Gymnasium tasks (e.g. classic control, MuJoCo, Atari) which may be classified as short-horizon and non-adversarial, the CTF Gymnasium task is a long-horizon adversarial game with a real-world counterpart for testing, making it an excellent focus problem for this work. This Python-based environment is a lightweight implementation of MOOS-IVP software that simulates CTF games between teams of unmanned surface vehicles \cite{moos-ivp}. As shown in Fig. \ref{fig:2}, the goal of the CTF task is to capture the opponent's flag by grabbing it and bringing it back to the team's side without getting tagged by an opponent. An agent will ``grab" the opponent flag when it comes within a certain distance of the flag while not in a tagged state. An agent becomes ``tagged'' when it comes within a certain distance of an opponent and both the agent and opponent are on the opponent's side of the environment. A tagged agent is forced to drive directly back to its home base (the flag region) and cannot participate in the game until it arrives there. To prevent agents from constantly guarding the flag by making consecutive tags, a tagging cooldown restricts each agent from tagging until a certain amount of time has passed since it last made a tag. 

The game is won by the team that makes the most flag captures within the time limit. In accordance with this, we define the score function as $\eta(\tau) = \text{Team Captures} - \text{Opponent Captures}$. Agents in this environment are able to observe the relative distance and heading to all other agents, flags, and boundaries; the tagging, on-sides, and flag grab status of each agent; and the number of captures made by each team. The environment returns this information as a vector with entries normalized between $[-1, 1]$. The low-level action space of agents consisted of 4 actions: do nothing, move forward at maximum speed, steer towards a relative heading of $-135^\circ
$ (steer right) at maximum speed, and steer towards a relative heading of $135^\circ
$ at maximum speed (steer left). In our experiments, we used a team size of $2$, a playing field size of $160 \times 80 \mathrm{m}^2$, a tagging and flag grab radius of $10\mathrm{m}$, a maximum agent speed of $1.5\mathrm{m}/\mathrm{s}$, a tagging cooldown of $30 \mathrm{s}$, a simulation update rate of $10\mathrm{Hz}$, and a simulation real time factor of $1/3$.

\subsection{Options-Level Behavioral Cloning}
\begin{figure}[ht]
    \vspace{2mm}
    \centering
    \includegraphics[width=\linewidth]{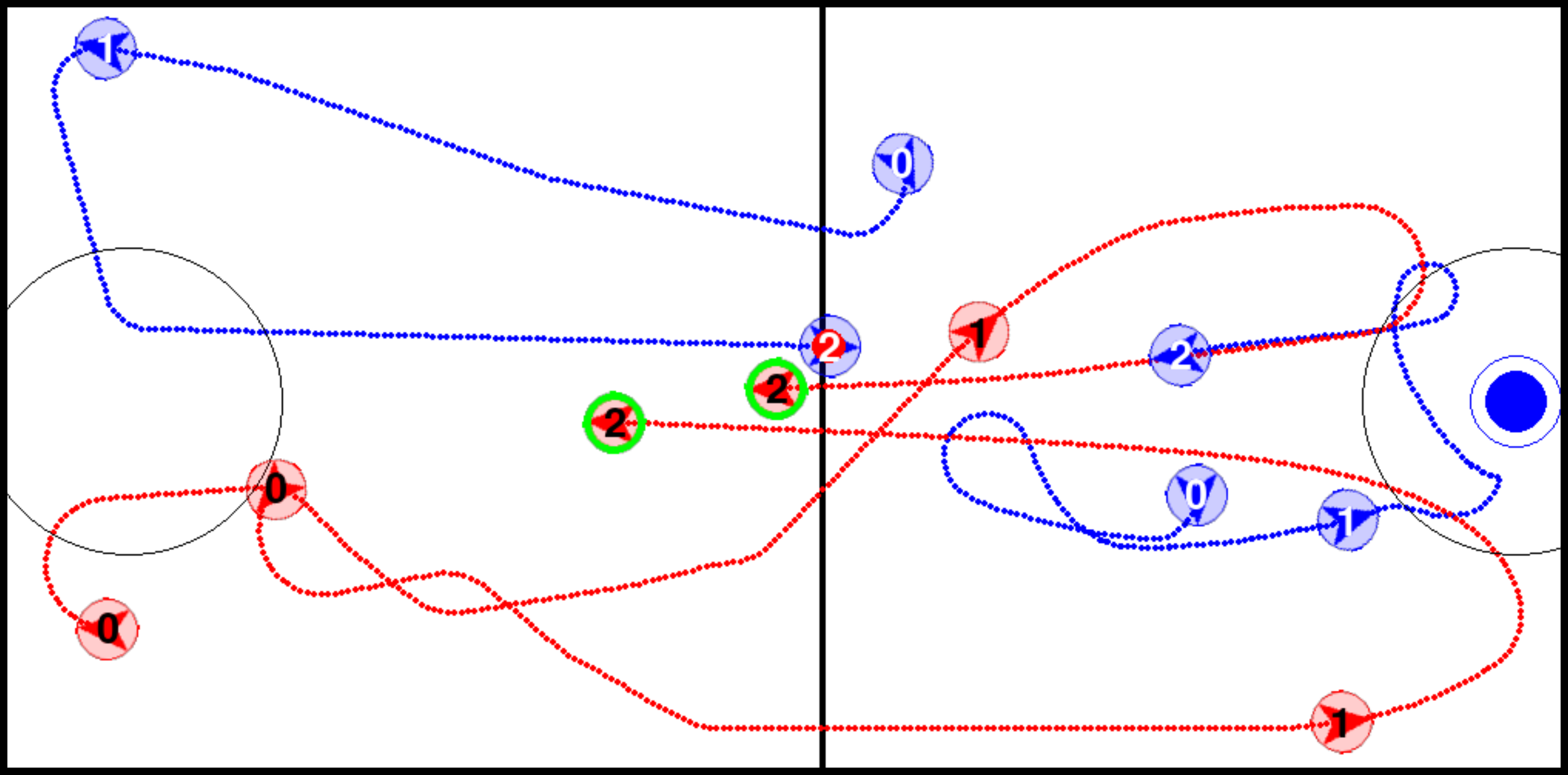}
    \caption{
        A CTF game played out in the Python simulator where the numbers on agents indicate points in time. The central vertical line divides the arena into a blue side (right) and a red side (left), and the large circles on the right and left ends represent the blue and red flag grab zones respectively. The blue team is using our options-BC policy and the red team is running a heuristic baseline. In this game, one member of the blue team captures the red flag while the other blue agent tags both red agents as they attempt to retrieve the blue flag. The green halo around the red agents at time 2 indicates that they are in a tagged state.
    }
    \label{fig:2}
\end{figure}

To perform options-level behavioral cloning (options-BC), we first pre-defined the intra-option policies $\pi_\omega$ with the following behavioral primitives: go to opponent flag region (``attack"), flank opponent flag region (``flank"), avoid opponents (``avoid"), retreat to own side (``retreat"), guard against opponents grabbing and capturing flag (``guard"), tag opponent closest to flag (``tag"), and do nothing (``no-op"). Each of these options is executed by controlling towards a desired speed and heading with the gymnasium environment's built-in PID controller.

To collect demonstrations at the options level, we constructed a wrapper class around the gymnasium environment that allows humans to control agents at the options level. For each agent, the wrapper class assigns each option to a unique key on a keyboard. When a demonstrator presses a key corresponding to option $\omega$ for a specific agent, that agent will follow $\pi_\omega$ until a different option key is pressed. Using this wrapper class, we collected 50 demonstrations in which a human played against agents running the \texttt{Heuristic\_CTF\_Agent} baseline. This policy is provided by the environment repository \cite{pyquaticus} and uses potential fields to attack and defend. For these demonstrations, we set the maximum time to $720\mathrm{s}$, and the maximum number of captures-per-team to $2$. Out of the 50 demonstrations, 47 were successful, meaning $\eta(\tau) > 0$.

We combined the option-level trajectories of each agent on the human team into a single dataset $\mathcal{D}_\Omega$ for training a shared policy-over-options $\pi_{\Omega_\text{BC}}$. We define $\pi_{\Omega_\text{BC}}$ as a fully-connected neural network with $2$ hidden layers of $256$ units with Tanh activations. The policy was trained on $\mathcal{D}_\Omega$ for 10 epochs using behavioral cloning with a learning rate of $0.001$ and a batch size of $32$. Fig. \ref{fig:2} illustrates a rollout of $\pi_{\Omega_\textbf{BC}}$ that shows the overall strategy learned from the demonstrations. Specifically, between time points 0 and 1, one blue agent executes the ``guard" option to stay between the opponents and its flag, while the other blue agent executes the ``flank" option to move along the edge of the environment towards the nearest corner on the red side. The low offensive presence of the blue team encourages both red agents to leave their flag unguarded and attack. Between time points 1 and 2, the blue defender uses the ``tag" option to tag the red agent closest to the flag and then attempts to tag the second red agent. If the second red agent is able to grab the flag before getting tagged, the blue defender will switch back to the ``guard" option to place itself in-between the second red agent and the scrimmage line. With both red agents either tagged or trapped on the blue side, the blue attacker can execute the ``attack" option to grab the flag and then the ``retreat" option to bring it back to the blue side.

To demonstrate the sample efficiency of options-level demonstrations, we also collected the low-level actions executed by the intra-option policies in the same set of demonstrations $\mathcal{D}_\Omega$ and trained a policy $\pi_\text{BC}(a|s)$ with BC on these state-action trajectories. We defined $\pi_\text{BC}$ with identical architecture to $\pi_{\Omega_\text{BC}}$ and followed the same training procedure. While the options-BC policy, $\pi_{\Omega_\text{BC}}$, won 99 out of 100 games against the baseline with an average score of 1.89, the vanilla BC policy won only 2 out of 100 games against the same baseline with an average score of -1.86.

\subsection{Reward Extrapolation}
We test SPLASH with the trained options-BC policy $\pi_{\Omega_\text{BC}}$ and the following hyperparameters: noise schedule $\mathcal{E} = (0.5, 0.67, 0.83, 1.0)$, $M=100$ rollouts per $\epsilon \in \mathcal{E}$, $\lambda_1=10$, $\lambda_2=20$, $\lambda_\text{IF}=4$, and an $80\%$ training $20\%$ validation split. Additionally, we set the maximum captures-per-team to 10, the maximum rollout time to $750\mathrm{s}$, and the trajectory downsampling rate to $40$ timesteps per sample. As in D-REX, we generate a set of no-op trajectories which are ranked beneath $\epsilon = 1.0$ to select for reward functions that prefer taking action over doing nothing. We withhold noise levels less than $0.5$ and discard trajectories with a score higher than 2 in order to evaluate SPLASH's ability to extrapolate beyond suboptimal demonstrations. To compare against the state-of-the-art, we test D-REX with the same options-BC policy, noise schedule, number of rollouts per $\epsilon$, and number of training trajectory pairs ($256,000$ for an $80$-$20$ training-validation split with 2 agents-per-team). In our experiments, we define $R_\theta$ as a fully-connected neural network with $2$ hidden layers of $256$ units with ReLU activations. We trained the SPLASH and D-REX reward functions for 10 epochs using the Adam optimizer with a learning rate of $1e-5$ and a batch size of $32$.

\begin{figure}[htb]
    \centering
    \begin{subfigure}{.49\columnwidth}
        \centering
        \includegraphics[width=\columnwidth]{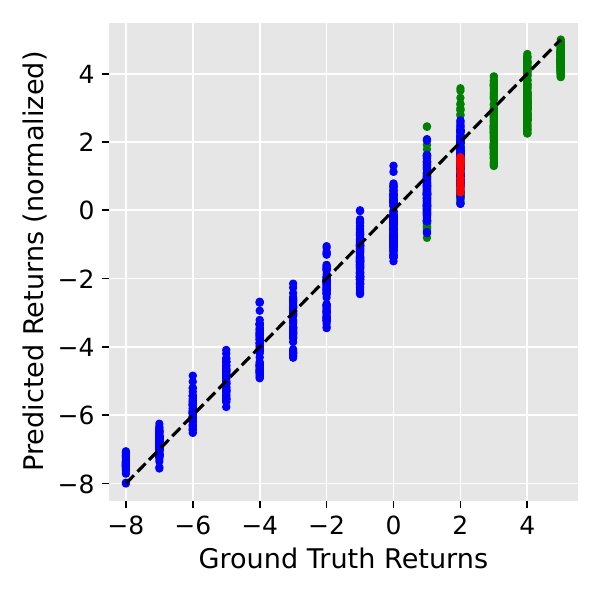}
        \caption{SPLASH}
        \label{fig:3a}
    \end{subfigure}
    \begin{subfigure}{.49\columnwidth}
        \centering
        \includegraphics[width=\columnwidth]{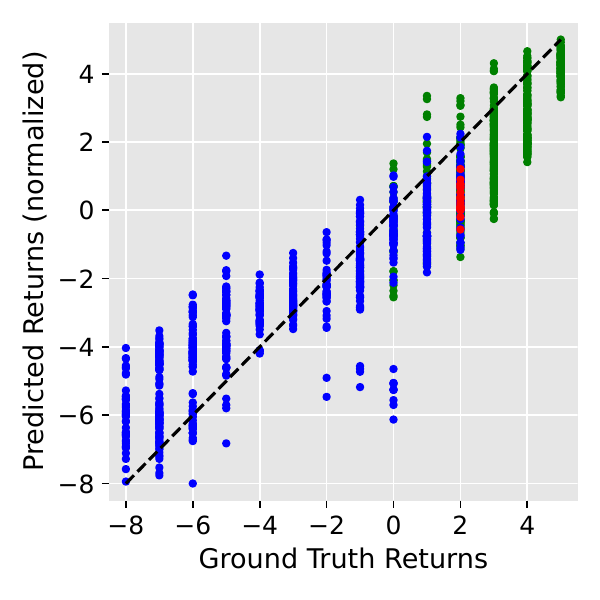}
        \caption{D-REX}
        \label{fig:3b}
    \end{subfigure}
    \caption{
        Reward extrapolation plots comparing SPLASH and D-REX predicted returns to ground truth for the CTF task. We normalize the predicted returns to fall within the range of the ground truth returns. Blue dots represent training trajectories generated by rolling out the options-BC policy at each noise level in the schedule. Red dots represent human demonstrations, and green dots represent previously unseen trajectories generated by rolling out the options-BC policy at lower noise levels than those in the schedule.}
    \label{fig:3}
\end{figure}

In Fig. \ref{fig:3}, we compare the alignment of SPLASH and D-REX to the ground truth reward. Predicted returns are calculated by summing the rewards $R_\theta(s)$ over a trajectory and normalizing by trajectory length. Across all bins, SPLASH exhibits significantly lower deviation from the ground truth compared to D-REX. Fig. \ref{fig:3} also shows the ability of SPLASH to accurately extrapolate beyond performance levels seen in the training trajectories and demonstrations.

\begin{figure}[b]
    \centering
    \includegraphics[width=\linewidth]{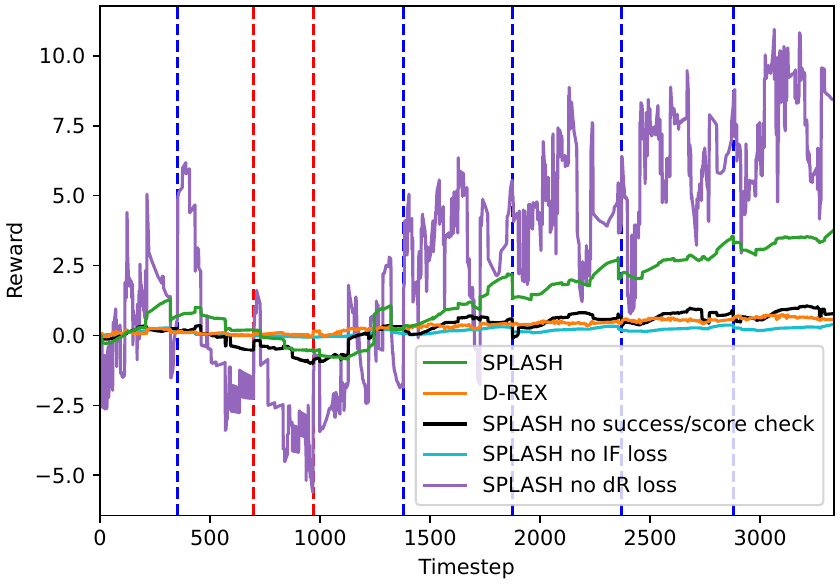}
    \caption{ 
        Blue team reward signals of SPLASH and D-REX for a game of CTF. Vertical blue and red lines indicate captures by the blue and the red teams, respectively. SPLASH is ablated by individually removing the score and success-based trajectory pair pruning, initial-final state comparisons, and reward derivative regularization. Reward signals are shifted to originate from 0 for easier comparison.
    }
    \label{fig:4}
\end{figure}

To test the ability of SPLASH to track task performance and progress, we plot the blue team's reward over time in a game of CTF. As shown in Fig. \ref{fig:4}, SPLASH is able to track both progress as the blue score increases, and loss of progress when the red score increases. In contrast, D-REX produces a reward curve with low signal-to-noise ratio that does not clearly indicate gains and losses of progress. To experimentally validate all components of SPLASH (success and progress-based learning constraints, temporal consistency regularization), we perform an ablation study in which SPLASH is run with individual components removed. As shown in Fig. \ref{fig:4}, these components help to denoise the reward signal and improve progress tracking.

\subsection{Hardware Experiments}
In table \ref{tab1}, we report results from a set of 3 CTF games played by Clearpath Heron uncrewed surface vehicles on the Charles River. The MOOS-IvP autonomy framework \cite{moos-ivp} is used on each vehicle to execute the agent policy while enforcing the game rules and maintaining safety. The python-implemented agent policies and MOOS-IvP run on a Raspberry Pi 4, the backseat computer in the frontseat-backseat architecture. Each vehicle can communicate with its teammates using 2.4 GHz radios and all messages are routed through a shoreside antenna. Game management apps in a shoreside MOOS community are used to arbitrate tag requests, flag grabs, and flag scores, while also communicating the full state of the game; including vehicle position, heading, and score; to all agents.

\begin{table}[tb]
\vspace{2mm}
\caption{
    Charles River CTF Experimental Results. The game events describe actions performed by each team. Starred numbers denote the best performance in these categories for each game.\vspace{-2mm}
}
\begin{center}
\resizebox{0.75\columnwidth}{!}{%
    \begin{tabular}{|c|c|cccc|}
    \hline
    \textbf{Game} &
      \textbf{Game} &
      \multicolumn{4}{c|}{\textbf{Game Events}} \\ \cline{3-6} 
    \textbf{Number} &
      \textbf{Duration} &
      \multicolumn{2}{c|}{\textit{\textbf{Captures}}} &
      \multicolumn{2}{c|}{\textit{\textbf{Grabs}}} \\ \cline{3-6} 
     &
      (seconds) &
      \multicolumn{1}{c|}{Blue} &
      \multicolumn{1}{c|}{Red} &
      \multicolumn{1}{c|}{Blue} &
      \multicolumn{1}{c|}{Red} \\ \hline
    1 &
      848 &
      \multicolumn{1}{c|}{1} &
      \multicolumn{1}{c|}{\textbf{\ 2$^*$}} &
      \multicolumn{1}{c|}{2} &
      \multicolumn{1}{c|}{\textbf{\ 4$^*$}} \\ \hline
    2 &
      570 &
      \multicolumn{1}{c|}{\textbf{\ 2$^*$}} &
      \multicolumn{1}{c|}{1} &
      \multicolumn{1}{c|}{2} &
      \multicolumn{1}{c|}{2} \\ \hline
    3 &
      891 &
      \multicolumn{1}{c|}{2} &
      \multicolumn{1}{c|}{2} &
      \multicolumn{1}{c|}{3} &
      \multicolumn{1}{c|}{3} \\ \hline
    \end{tabular}%
}
\label{tab1}
\end{center}
\end{table}

As in the simulated experiments, we run games between the options-BC policy from SPLASH (blue team) and the heuristic baseline from the gymnasium repository (red team). These experiments demonstrate the robustness of our approach to deviations from the idealized environment in the Python simulator. During these games the vehicles had to contend with naturally occurring disturbances such as time-varying winds, and performance was constrained by hardware limitations such as finite actuator bandwidth, impacting the ability to execute maneuvers. Table \ref{tab1} displays the results of the 3 experiments. The 1 loss, 1 win and 1 tie record of the options-BC policy (blue) shows that it was able to compete with the potential-based heuristic (red).

\begin{figure}[!h]
    \centering
    \includegraphics[width=0.9\linewidth]{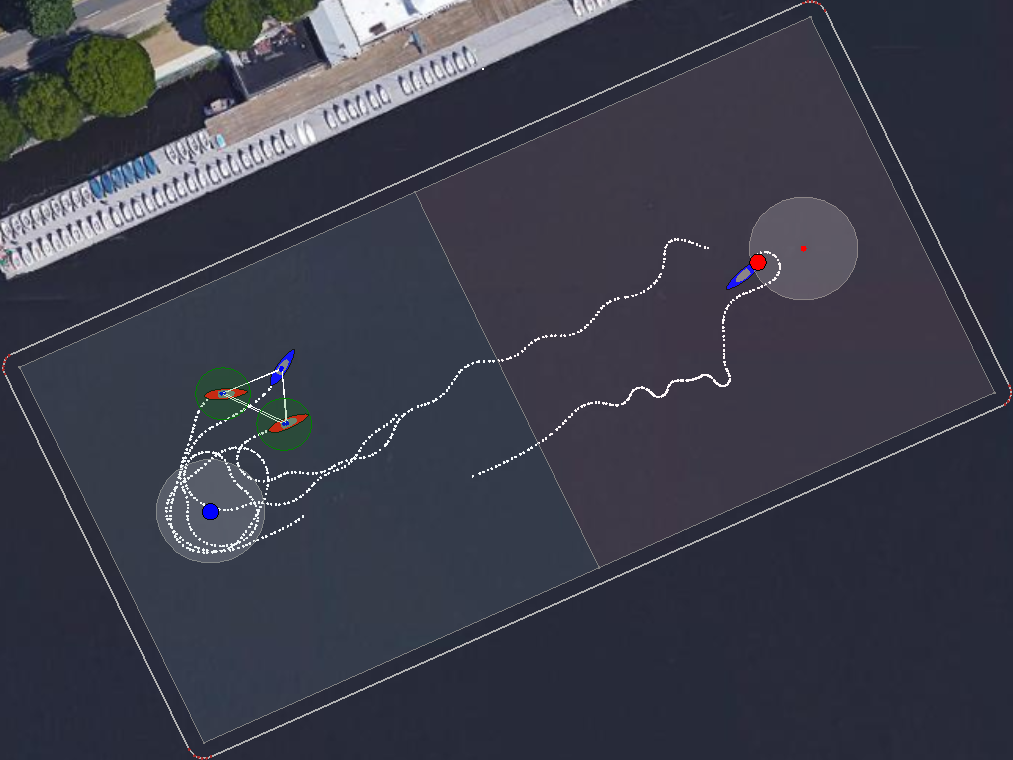}
    \caption{
        GPS visualization of real-world Heron surface vehicles 824 seconds into game number 3 of the Charles River experiments. The blue team is running our options-BC policy and the red team is running a potential-based heuristic baseline from the CTF gymnasium repository \cite{pyquaticus}.
    }
    \label{fig:5}
\end{figure}

As shown in Fig. \ref{fig:5}, the options-BC policy executed the same high-level strategy observed in the gymnasium environment: one agent defends the team flag by tagging and trapping the attacking opponents, while the other retrieves the opponents' flag. On multiple occasions, we observed a blue attacker getting pushed out of bounds by the current as it attempted to move along the edge of the playing field. This would force the agent to return to its home base before re-joining the game. We suspect that more captures would have been made by the blue team if this did not occur. Overall, these experiments demonstrate the sim-to-real translatability of the options-BC policy, which in turn shows that SPLASH can learn accurate reward functions for tasks with real-world complexity.

\section{Conclusion} \label{sec:conclusion}
In this work, we introduce SPLASH, a preference-based IRL algorithm for learning from suboptimal demonstrations of long-horizon and adversarial tasks. We empirically validated the ability of SPLASH to learn accurate reward functions on a maritime CTF task and demonstrated the sim-to-real translatability of an options-level policy in this domain. In the future, we plan to modify SPLASH to handle states comprised of a game history and then test SPLASH reward functions on state-of-the art RL algorithms that utilize recurrent neural networks and intelligent exploration.

\section{Acknowledgment} \label{sec:acknowledgment}
This work was partially supported by NSF 2219101 and AFOSR FA9550-23-1-0529 at Boston University. We thank Christopher LeMay for providing demonstrations in the CTF gymnasium environment for this research.







\bibliographystyle{IEEEtran}

\end{document}